\documentclass{article}

\PassOptionsToPackage{numbers, compress}{natbib}
\usepackage[preprint]{neurips_2026}

\usepackage[utf8]{inputenc}
\usepackage[T1]{fontenc}
\usepackage{hyperref}
\usepackage{url}
\usepackage{booktabs}
\usepackage{amsmath}
\usepackage{comment}
\usepackage{amssymb}
\usepackage{amsthm}
\usepackage{amsfonts}
\usepackage{nicefrac}
\usepackage{microtype}
\usepackage{xcolor}
\usepackage{multirow}
\usepackage{graphicx}
\usepackage{algorithm}
\usepackage{float}
\usepackage{wrapfig}
\usepackage{listings}
\usepackage{fontawesome5}

\definecolor{codeblue}{rgb}{0.25, 0.5, 0.5}
\definecolor{codekw}{rgb}{0.85, 0.18, 0.50}
\definecolor{codesign}{RGB}{0, 0, 255}
\usepackage{algpseudocode}
\renewcommand{\algorithmicrequire}{\textbf{Input:}}
\usepackage{enumitem}
\hypersetup{
  colorlinks=true,
  linkcolor={red},
  citecolor={orange!70!black},
  urlcolor={blue!80!black},
}
\definecolor{spblue}{HTML}{00b5ea}
\definecolor{spred}{HTML}{E74C3C}
\definecolor{spgreen}{HTML}{2ECC71}
\definecolor{lightblue}{rgb}{0.88, 0.95, 0.98}
\definecolor{lightpink}{rgb}{0.98, 0.88, 0.95}
\definecolor{superlightred}{rgb}{0.99, 0.92, 0.92}
\definecolor{lightred}{rgb}{0.98, 0.88, 0.88}

\newcommand{\qpsi}{q_{\psi}}
\newcommand{\ptgt}{p_{\theta}}
\title{D-PACE: Dynamic Position-Aware Cross-Entropy for Parallel Speculative Drafting}

\author{
    Tianyu Wu\textsuperscript{1}
    \thanks{\textbf{\textcolor{red}{Project Lead.}} Correspondence to \texttt{lucaswu@fas.harvard.edu}.}
    \hspace{0.1em}
    \thanks{Equal contribution.}
    \quad
    Yu Yao\textsuperscript{2}\footnotemark[2]
    \quad
    Zhenting Qi\textsuperscript{1}
    \quad
    Han Zheng\textsuperscript{2}
    \quad
    Zhuohan Wang\textsuperscript{1}
    \\
    \textbf{
        Haoran Ma\textsuperscript{1}
        \quad
        Lawrence Liao\textsuperscript{1}
        \quad
        Himabindu Lakkaraju\textsuperscript{1}
        \quad
        Ju Li\textsuperscript{2}\thanks{Equal advising.}
        \quad
        Yilun Du\textsuperscript{1}\footnotemark[3]
    }
    \\
    [0.1cm]
    \textsuperscript{1}Harvard
    \quad
    \textsuperscript{2}MIT
    \\
    [0.1cm]
    {
        \href{https://github.com/Lucas-TY/D-PACE}{\textcolor{black}{\faGithub}~https://github.com/Lucas-TY/D-PACE}
    }
}

\begin{document}

\maketitle

\vspace{-20pt}
\begin{abstract}

Speculative decoding accelerates LLM inference by having a small drafter propose tokens that a larger target model verifies in parallel. Recent diffusion-based parallel drafters such as DFlash predict the full $B$-token block in one forward pass, enabling deeper drafters and longer accepted blocks. However, existing multi-token drafter objectives often use fixed position-dependent weighting schedules, such as head-dependent weights or block-position decays, which do not adapt as the positions limiting acceptance change during training. To address this, we derive per-position training weights from a differentiable surrogate of expected accepted draft length, matching the weight of each position to its log-probability gradient contribution. The resulting loss, \textbf{D-PACE (Dynamic Position-Aware Cross-Entropy)}, shifts training signal toward positions that currently limit acceptance as the drafter improves. Across six benchmarks, two Qwen3-4B draft depths, two decoding temperatures, and two additional target models, D-PACE consistently improves both wall-clock speedup and average emitted length, with 2.3\% measured training-time overhead and no changes to the drafter architecture or inference procedure.

\end{abstract}

\section{Introduction}
\label{sec:intro}

Autoregressive decoding is the dominant latency bottleneck of LLM inference because token-by-token generation exposes limited parallelism and often leaves modern GPUs underutilized, especially for the long chains of thought produced by reasoning models~\citep{achiam2023gpt4,touvron2023llama,touvron2023llama2,guo2025deepseekr1}. To reduce this sequential bottleneck, speculative decoding (SD)~\citep{leviathan2023speculative,chen2023speculative,xia2024survey} uses a small draft model to propose tokens that a larger target model verifies in parallel.

\emph{Autoregressive drafters with tree decoding}, exemplified by EAGLE~\citep{li2024eagle,li2024eagle2,li2025eagle3,zhang2025hass}, draft tokens one at a time at inference, with the drafter autoregressively continuing on its own previous outputs. This sequential drafting process limits practical speedups and often constrains practical drafters to shallow architectures under latency budgets~\citep{chen2026dflash}. \emph{Parallel single-sequence} methods, building on block-diffusion language models~\citep{arriola2025block,nie2025llada} and exemplified by DFlash~\citep{chen2026dflash}, instead generate all $B$ tokens in a single forward pass, reducing the sequential dependence on block length and allowing deeper, more expressive draft models within the same latency budget.

Parallel block drafting reduces draft-generation latency, but acceptance remains prefix-based: an error at an early position rejects the following draft tokens. Existing objectives therefore often use position-dependent weights. For example, Medusa~\citep{cai2024medusa} uses head-dependent loss weights, while DFlash~\citep{chen2026dflash} applies a fixed exponential decay within each draft block. These schedules encode prefix importance but depend only on position, rather than on which positions currently limit acceptance for a given drafter and example. A fixed schedule therefore misallocates training signal: when early positions are already highly accurate, later positions become the acceptance bottleneck and should receive more weight; when an earlier prediction is incorrect, later positions cannot be accepted regardless of their quality, so training signal assigned to them cannot improve the realized accepted prefix.

\begin{figure}[t]
\centering
\includegraphics[width=\linewidth]{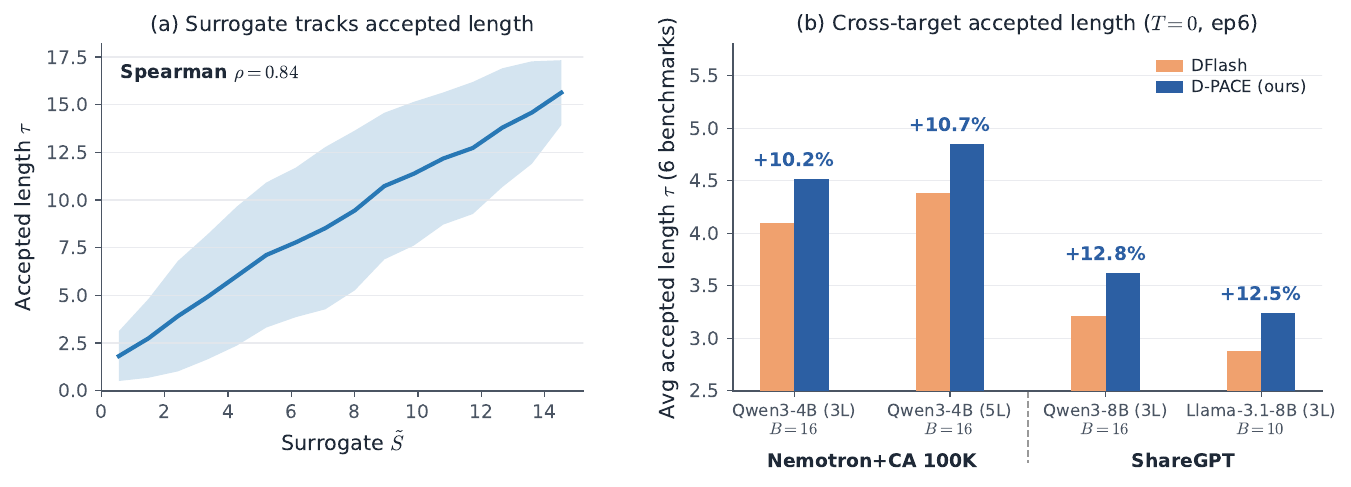}
\vspace{-0.4em}
\caption{\textbf{Surrogate $\tilde S$ tracks emitted length $\tau$; D-PACE improves $\tau$ across targets.} \textbf{(a)} Per-block surrogate $\tilde S = \sum_{k=1}^{B}\prod_{i\le k} q_i$ ($q_i$: draft confidence on the token selected by the target decoding policy at position $i$) versus reported $\tau$, computed during decoding on the 3L DFlash baseline over 128 MATH-500 prompts; line: bin means; shaded band: $\pm 1\sigma$. Spearman $\rho{=}0.84$ (App.~\ref{app:correlation}). \textbf{(b)} Average emitted length $\tau$ over six benchmarks at $T{=}0$ (epoch 6) for the DFlash baseline and D-PACE across four configurations: Qwen3-4B (3L and 5L, $B{=}16$), Qwen3-8B (3L, $B{=}16$), and Llama-3.1-8B (3L, $B{=}10$). Relative improvement is labeled above each pair.}
\label{fig:motivation}
\end{figure}

To replace the fixed schedule, we seek a training signal that reflects each position's contribution to accepted length. Accepted length is the natural inference-time target, but per-position acceptance is a discrete prefix event determined by the verifier. EAGLE-2~\citep{li2024eagle2} observes at inference that the cumulative product of draft confidences along a path correlates with that path's acceptance rate. In drafters without tree decoding such as DFlash, there is only one such path per block, so this cumulative confidence can serve as a smooth surrogate of the block's expected accepted draft length during training; differentiating it yields each position's contribution under this accepted-length proxy. We empirically verify this on the 3L DFlash baseline (MATH-500): the smooth surrogate $\tilde S$ (formally defined in Sec.~\ref{sec:surrogate}) is strongly correlated with reported $\tau$ (Fig.~\ref{fig:motivation}; App.~\ref{app:correlation}). We use these contributions as per-position weights in our weighted cross-entropy loss, which we call \textbf{D-PACE} (Dynamic Position-Aware Cross-Entropy).

\paragraph{Contributions.} \textbf{Adaptive per-position weights.}
We derive a weighted cross-entropy objective from a smooth surrogate of expected accepted draft length for parallel single-trajectory drafters. The derivation yields closed-form per-position weights that adapt to which block positions currently limit acceptance for the given drafter and example, replacing DFlash's fixed exponential schedule without changing the drafter architecture or decoding procedure. \textbf{Stable training via asymmetric smoothing.}
D-PACE stabilizes the cumulative-product weights with asymmetric smoothing: smoothing is applied only when computing the weights, while the cross-entropy term remains on the unsmoothed target-token probability. This preserves token-level CE gradients while preventing early low-confidence positions from suppressing later-position weights. \textbf{Evaluation across settings.}
On Qwen3-4B, D-PACE improves average decoding speedup by $+8.0\%$--$9.7\%$ and average emitted length by $+8.5\%$--$10.7\%$ across the main Qwen3-4B configurations (Table~\ref{tab:main}). On Llama-3.1-8B-Instruct and Qwen3-8B, D-PACE improves average emitted length by about $+12.5\%$ and $+12.8\%$, respectively (Table~\ref{tab:cross-model}).

\section{Preliminaries}
\label{sec:prelim}

\paragraph{Speculative decoding notation.}
Let $p_\theta$ denote a large \emph{target} model and $q_\psi$ a smaller \emph{draft} model. At each decoding step, the draft generates a block of $B$ candidate tokens $z_1,\ldots,z_B$ from $q_\psi$, and the target verifies all $B$ tokens in a single forward pass. Verification proceeds left to right: the first position $j$ at which the draft token is rejected terminates the block, and all tokens $z_1,\ldots,z_{j-1}$ before it are accepted. The number of accepted draft tokens per block is the \emph{accepted draft length}
\begin{equation}
X \;=\; \max\bigl\{k \in \{0,\ldots,B\} : z_i \text{ accepted for all } i \le k\bigr\},
\label{eq:acc-len}
\end{equation}
and, for fixed draft and verification costs, the expected accepted draft length $\mathbb{E}[X]$ is a key factor in the speedup over standard autoregressive decoding~\citep{leviathan2023speculative,chen2023speculative}.

\paragraph{DFlash block drafter and baseline objective.}
DFlash~\citep{chen2026dflash} is the parallel block drafter used in our main experiments. Given a verified context, it predicts all $B$ draft positions in one forward pass using a block decoder with KV injection from the target's prefill cache. Let $z_i^*$ denote the token selected by the target decoding policy at block position $i$ and $q_i := q_\psi(z_i^* \mid \text{prefix})$ the draft confidence on that token. The DFlash decayed-CE baseline trains the drafter with
\begin{equation}
\mathcal{L}_{\text{DFlash}} \;=\; \sum_{j=1}^{B}\exp\!\Bigl(-\frac{j-1}{\gamma}\Bigr)\,(-\log q_j).
\label{eq:dflash-loss}
\end{equation}
These fixed coefficients define the baseline that D-PACE replaces in Sec.~\ref{sec:method}.

\vspace{-5pt}
\section{D-PACE: Dynamic Position-Aware Cross-Entropy}
\label{sec:method}

\begin{wrapfigure}{r}{0.44\linewidth}
\vspace{-30pt}
\begin{minipage}{\linewidth}
\begin{algorithm}[H]
\caption{D-PACE Optimization}
\label{alg:dpace}
\begin{lstlisting}[basicstyle=\ttfamily\scriptsize]
Input: draft confidences
       $q_i = q_\psi(z_i^* \mid \mathrm{prefix})$

for $i$ = 1, ..., $B$:
    # smoothing
    $\tilde{q}_i$ = $(1 - \alpha) \cdot q_i + \alpha$  
    
for $j$ = 1, ..., $B$:
    # prefix products
    $P_j$ = $\prod_{i=1}^{j} \tilde{q}_i$  
    
for $j$ = 1, ..., $B$:
    # suffix sums of prefix products
    $\bar{w}_j$ = $\operatorname{sg}(\sum_{m=j}^{B} P_m)$  
    
return $\sum_{j=1}^{B}\bar{w}_j\cdot(-\log q_j)$
\end{lstlisting}
\end{algorithm}
\end{minipage}
\vspace{-25pt}
\end{wrapfigure}

\vspace{-10pt}
\subsection{Accepted-Length Surrogate}
\label{sec:surrogate}
Using the notation from Sec.~\ref{sec:prelim}, let $a_i$ be the per-position acceptance rate, i.e., the probability that position $i$ is accepted given that the first $i{-}1$ positions are accepted. By the chain rule of probability, $\mathbb{P}(X\ge k) = \prod_{i\le k} a_i$. Combined with the tail-sum identity $\mathbb{E}[X] = \sum_{k\ge 1}\mathbb{P}(X\ge k)$ for non-negative integer $X$, this gives
\begin{equation}
\mathbb{E}[X] \;=\; \sum_{k=1}^{B}\mathbb{P}(X\ge k) \;=\; \sum_{k=1}^{B}\prod_{i=1}^{k}a_i.
\label{eq:E-acc}
\end{equation}
The acceptance rates $\{a_i\}$ in Eq.~\eqref{eq:E-acc} are not available during training; we use the draft confidence $q_i$ as a smooth proxy for $a_i$ (supported by Fig.~\ref{fig:motivation}) to obtain the closed-form surrogate
\begin{equation}
\mathbb{E}[X] \;\approx\; \tilde S(\psi) \;:=\; \sum_{k=1}^{B}\prod_{i=1}^{k}q_i,
\label{eq:S}
\end{equation}
which we use as a smooth proxy for deriving per-position training weights.

\begin{figure}[t]
\centering
\includegraphics[trim=28pt 0 28pt 0,clip,width=\linewidth]{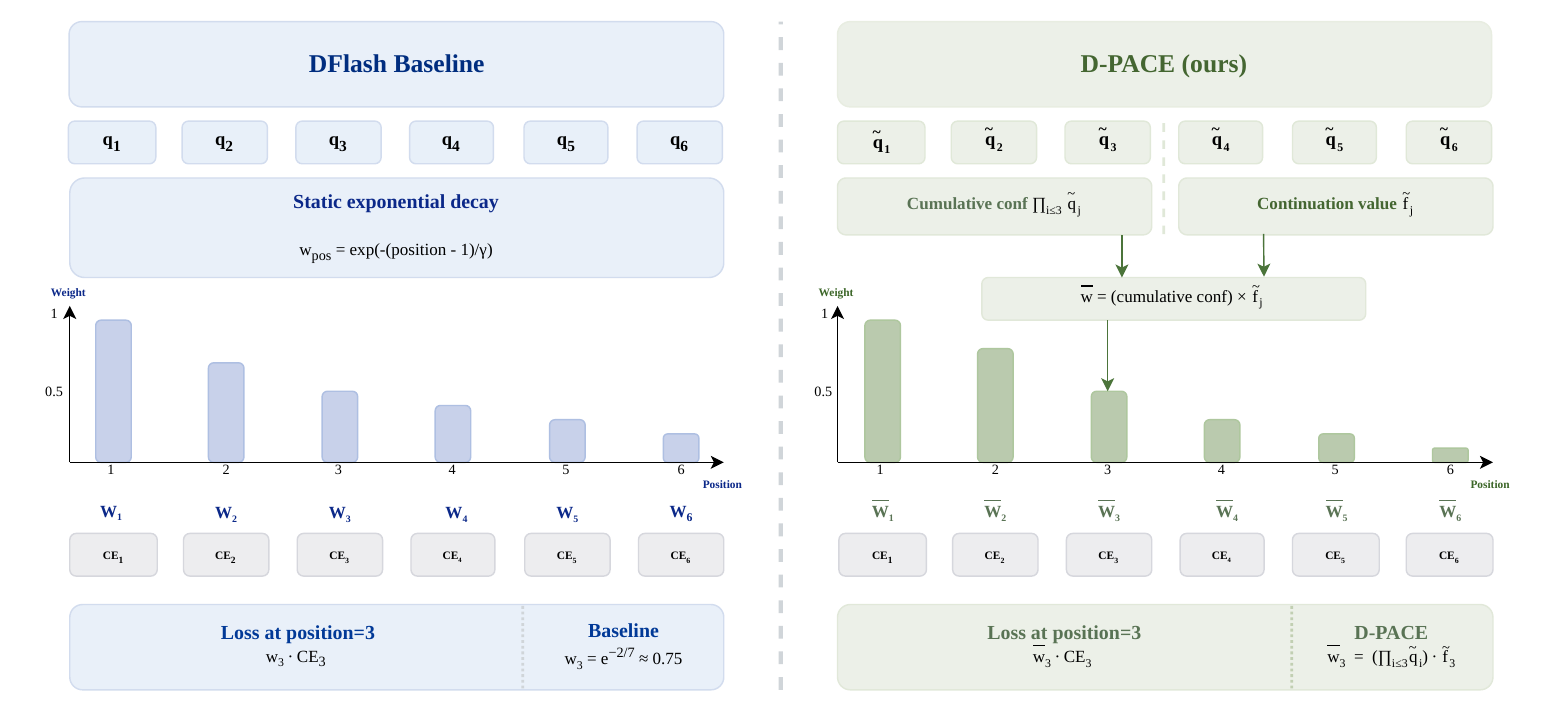}
\vspace{-0.4em}
\caption{\textbf{D-PACE versus the DFlash baseline.} \textbf{Left}: DFlash applies a fixed exponential position decay. \textbf{Right}: D-PACE uses example-dependent weights that combine prefix confidence with remaining accepted-length value. Both are weighted cross-entropy objectives with different position weights.}
\label{fig:dpace-example}
\end{figure}

Fig.~\ref{fig:dpace-example} illustrates the difference between DFlash's fixed decay and D-PACE's adaptive weights.

\subsection{Per-Position Coefficient}
\label{sec:coefficient}

To obtain a per-position weight, we differentiate the block-level surrogate $\tilde S$ with respect to each $q_j$. For any $j \in \{1,\ldots,B\}$, we split the outer sum in Eq.~\eqref{eq:S} at $k{=}j$ into terms with and without $q_j$:
\begin{equation}
\tilde S(\psi) \;=\; \sum_{k=1}^{j-1}\prod_{i=1}^{k}q_i \;+\; \prod_{i<j} q_i \cdot q_j \cdot \Bigl(1 + \sum_{m=j+1}^{B}\prod_{i=j+1}^{m} q_i\Bigr) \;=\; \sum_{k=1}^{j-1}\prod_{i=1}^{k}q_i \;+\; \Bigl(\prod_{i<j} q_i\Bigr) q_j\, f_j,
\end{equation}
where
$
f_j \;:=\; 1 + \sum_{m=j+1}^{B}\prod_{i=j+1}^{m} q_i
$
is the {continuation value} (the accepted-length value contributed by position $j$ and later positions, conditioned on accepting the prefix through $j$). The first sum is constant in $q_j$, so $\partial \tilde S/\partial q_j = (\prod_{i<j} q_i) f_j$: the contribution of position $j$ is modulated by the prefix-confidence factor $\prod_{i<j} q_i$. Applying the log-derivative identity $\nabla_\psi q_j = q_j \nabla_\psi \log q_j$ yields the closed-form gradient
\begin{equation}
\nabla_\psi \tilde S \;=\; \sum_{j=1}^{B} \Bigl(\prod_{i<j} q_i\Bigr) f_j\, \nabla_\psi q_j \;=\; \sum_{j=1}^{B} w_j\, \nabla_\psi \log q_j, \qquad w_j \;:=\; \Bigl(\prod_{i\le j} q_i\Bigr) f_j.
\label{eq:grad}
\end{equation}
Since $w_j$ depends only on the draft confidences $\{q_i\}$ in Eq.~\eqref{eq:grad}, D-PACE is \emph{draft-only}: after the target-generated training tokens are available, computing the weights requires no additional target-model forward pass. The weight $w_j$ splits into the {cumulative confidence} $\prod_{i\le j}q_i$ (a proxy for accepting the prefix through position $j$) and the {continuation value} $f_j$.

\subsection{Asymmetric Weight Smoothing}
\label{sec:smoothing}

The raw weight $w_j = (\prod_{i\le j} q_i) f_j$ is numerically unstable: its cumulative factor $\prod_{i\le j} q_i$ can vanish when multiple small values of $q_i$ are multiplied across positions. To prevent this, we smooth each $q_i$ inside the weight by interpolating toward 1 with mixing parameter $\alpha \in [0,1]$:
\begin{equation}
\tilde q_i \;:=\; (1-\alpha)\, q_i + \alpha.
\label{eq:smoothing}
\end{equation}
We apply the smoothing in Eq.~\eqref{eq:smoothing} only inside the weight, leaving $-\log q_j$ unsmoothed so that the cross-entropy retains its standard $1/q_j$ gradient. With $\tilde q_i \ge \alpha$, the cumulative product through position $j$ is at least $\alpha^j$, so later weights cannot vanish. Let $\tilde f_j$ denote the smoothed analog of $f_j$ (i.e., $f_j$ with $q_i \to \tilde q_i$). Applying the same substitution to every $q_i$ in $w_j = (\prod_{i\le j} q_i) f_j$ yields the per-position weight
\begin{equation}
\bar w_j \;:=\; \Bigl(\prod_{i\le j}\tilde q_i\Bigr)\, \tilde f_j \;=\; \Bigl(\prod_{i\le j}\tilde q_i\Bigr)\,\Bigl(1 + \sum_{m=j+1}^{B}\prod_{i=j+1}^{m}\tilde q_i\Bigr) \;=\; \sum_{m=j}^{B}\prod_{i\le m}\tilde q_i .
\label{eq:dpace-coef}
\end{equation}

Substituting $\bar w_j$ as the per-position weight in a weighted cross-entropy gives the D-PACE loss
\begin{equation}
\mathcal{L}_{\text{D-PACE}}(\psi) \;=\; \sum_{j=1}^{B} \bar w_j \,(-\log q_j),
\label{eq:dpace-loss}
\end{equation}
where $\bar w_j$ is detached from the gradient (treated as a constant). The weights therefore provide credit assignment for the cross-entropy terms without introducing gradients that decrease a position's own accepted-length contribution. Alg.~\ref{alg:dpace} computes the loss in Eq.~\eqref{eq:dpace-loss} from the smoothed confidences, their prefix products, and the corresponding suffix sums in Eq.~\eqref{eq:dpace-coef}.

\clearpage
\newpage

\section{Experiments}

\begin{figure}[h]
\centering
\includegraphics[width=\linewidth]{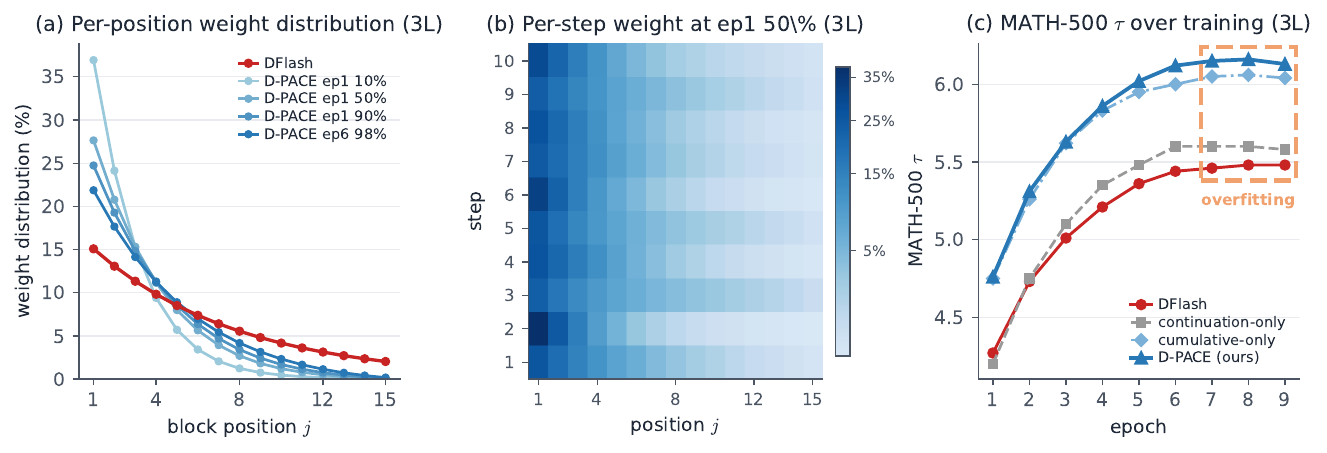}
\vspace{-0.4em}
\caption{\textbf{Weight dynamics and reported emitted length under D-PACE.} \textbf{(a)} On 3L DFlash drafts of Qwen3-4B, mean per-position weight averaged over 100 training steps at four checkpoints (10\%, 50\%, 90\% of epoch~1, and end of epoch~6): DFlash's exponential decay (red) versus D-PACE's per-position weights across checkpoints (blue). \textbf{(b)} Per-step weight over 10 consecutive training steps at epoch 1 (50\%) (Sec.~\ref{sec:weight-dynamics}). \textbf{(c)} MATH-500 reported $\tau$ over training (3L draft, $T{=}0$) for the DFlash baseline, D-PACE, and its two component-ablation variants (continuation-only, cumulative-only) (Sec.~\ref{sec:component-ablation}).}
\label{fig:teaser-results}
\end{figure}
In this section, we conduct experiments to address the following research questions:                                                          
 \begin{itemize}[leftmargin=*]
    \item \textbf{RQ1 (Efficacy):} How does D-PACE compare with related baselines?
   \item \textbf{RQ2 (Ablation):} How do D-PACE's cumulative-confidence and continuation-value factors affect performance?
   \item \textbf{RQ3 (Generalizability):} Does D-PACE transfer from Qwen3-4B to other target models?
   \item \textbf{RQ4 (Sensitivity):} How sensitive is D-PACE to smoothing and block-size choices?
  \end{itemize}

\subsection{Experimental Setup}
\label{sec:setup}

\noindent\textbf{Models and Benchmarks.}
For \textbf{Qwen3-4B}~\citep{yang2025qwen3}, we train 3L and 5L DFlash drafters, where 3L/5L denotes the number of draft Transformer layers. All reported drafters use three target hidden-state features. We evaluate the Qwen3-4B drafters at $T = 0$ and $T = 1$. For cross-model generalization (Sec.~\ref{sec:cross-model}), we additionally train 3L drafters for \textbf{Llama-3.1-8B-Instruct}~\citep{dubey2024llama3} and \textbf{Qwen3-8B}~\citep{yang2025qwen3}. We evaluate on six benchmarks: \textbf{Math} (GSM8K~\citep{cobbe2021gsm8k}, MATH-500~\citep{lightman2024verify,hendrycks2021math}), \textbf{Code} (HumanEval~\citep{chen2021humaneval}, MBPP~\citep{austin2021mbpp}), and \textbf{Chat} (MT-Bench~\citep{zheng2023mtbench}, Alpaca~\citep{alpaca}). For each benchmark, we report average emitted length $\tau$ and wall-clock speedup over autoregressive decoding (SR); higher is better for both. Training uses a single NVIDIA H200 GPU; SR is measured on an L40S. For Qwen3 targets, we disable thinking mode during evaluation, following DFlash.


\noindent\textbf{Datasets.}
For Qwen3-4B (main experiments), we use the {Qwen3-4B-Instruct-100K} dataset released by the DFlash authors~\citep{chen2026dflash}, whose data mixture is based on Nemotron-Post-Training-Dataset-v2~\citep{nathawani2025nemotron} and CodeAlpaca~\citep{chaudhary2023codealpaca}. The dataset provides prompts with target-generated responses for drafter training, and we use it directly to match the released DFlash training data setting. For Llama-3.1-8B-Instruct and Qwen3-8B, we use ShareGPT~\citep{sharegpt2023} and generate target-model responses for drafter training, following the EAGLE series~\citep{li2024eagle,li2024eagle2,li2025eagle3}.

\noindent\textbf{Implementation.}
Training is implemented in SpecForge~\citep{specforge2025}, the SGLang team's public speculative-decoding training framework, which provides a third-party implementation of DFlash. Evaluation uses the native benchmark code released by the DFlash authors with the same decoding settings for all methods. We follow DFlash's~\citep{chen2026dflash} block-size defaults for each target: $B{=}16$ for Qwen3-4B and Qwen3-8B, $B{=}10$ for Llama-3.1-8B. Training uses AdamW (learning rate $6{\times}10^{-4}$, effective batch size 4: micro-batch 2 $\times$ accumulation 2); full hyperparameters are in App.~\ref{app:training}. Unless otherwise stated, D-PACE uses smoothing $\alpha=0.5$ and results use the epoch-6 checkpoint.

\noindent\textbf{Baselines.}
We compare D-PACE against the \textbf{DFlash decayed-CE baseline} (fixed exponential position decay~\citep{chen2026dflash}) and \textbf{Top-3 prefix mask}, a hard 0/1 CE mask adapted from GRIFFIN~\citep{hu2025griffin}: the loss at position $j$ is included only when every earlier target token lies in the draft's top-3 predictions. We also include \textbf{Accept-rate}, which directly optimizes the accepted-length surrogate in Eq.~\ref{eq:S} and uses neither the detached weighted-CE construction nor asymmetric weight smoothing. We adopt only the GRIFFIN mask construction, not the full GRIFFIN method; all baselines are formalized in App.~\ref{app:losses}.

\subsection{Main Results}

For Qwen3-4B, D-PACE improves both SR and $\tau$ over the DFlash decayed-CE baseline for every setting reported in Table~\ref{tab:main}. On 3L MATH-500, D-PACE follows a similar convergence trajectory while maintaining a roughly $12\%$ higher $\tau$ than the baseline throughout the 9-epoch run (Fig.~\ref{fig:teaser-results}c).

\begin{table}[t]
\caption{\textbf{Main result.} Decoding speedup (SR) and average emitted length ($\tau$) on six benchmarks at two draft depths (3L, 5L) and two temperatures ($T{=}0,1$) for DFlash drafts of Qwen3-4B at epoch 6. ``Top-3 prefix mask'' is the hard-mask baseline adapted from GRIFFIN~\citep{hu2025griffin}. ``Accept-rate'' directly optimizes the accepted-length surrogate in Eq.~\ref{eq:S}; see Sec.~\ref{sec:setup} and App.~\ref{app:losses}.}
\label{tab:main}
\centering
\footnotesize
\setlength{\tabcolsep}{3.2pt}
\resizebox{\linewidth}{!}{%
\begin{tabular}{cl|cc|cc|cc|cc|cc|cc|cc}
\toprule
 & & \multicolumn{4}{c|}{\textsc{Math}} & \multicolumn{4}{c|}{\textsc{Code}} & \multicolumn{4}{c|}{\textsc{Chat}} & \multicolumn{2}{c}{} \\
\cmidrule(lr){3-6}\cmidrule(lr){7-10}\cmidrule(lr){11-14}
 & & \multicolumn{2}{c|}{GSM8K} & \multicolumn{2}{c|}{MATH-500} & \multicolumn{2}{c|}{HumanEval} & \multicolumn{2}{c|}{MBPP} & \multicolumn{2}{c|}{MT-Bench} & \multicolumn{2}{c|}{Alpaca} & \multicolumn{2}{c}{Avg.} \\
Depth & Method & SR & $\tau$ & SR & $\tau$ & SR & $\tau$ & SR & $\tau$ & SR & $\tau$ & SR & $\tau$ & SR & $\tau$ \\
\midrule
\multicolumn{16}{c}{\emph{Temperature} $T = 0$} \\
\midrule
\multirow{4}{*}{3L}
  & DFlash & 3.40 & 4.57 & 3.86 & 5.44 & 3.38 & 4.66 & 3.20 & 4.31 & 2.03 & 3.22 & 1.72 & 2.41 & 2.93 & 4.10 \\
  & Top-3 prefix mask & 3.29 & 4.44 & 3.67 & 5.19 & 3.22 & 4.43 & 3.06 & 4.13 & 1.95 & 3.11 & 1.62 & 2.29 & 2.80 & 3.93 \\
  & Accept-rate & 0.76 & 1.01 & 0.75 & 1.01 & 0.77 & 1.03 & 0.79 & 1.03 & 0.76 & 1.04 & 0.79 & 1.04 & 0.77 & 1.03 \\
  & \textbf{D-PACE} & \textbf{3.75} & \textbf{5.09} & \textbf{4.32} & \textbf{6.12} & \textbf{3.71} & \textbf{5.12} & \textbf{3.47} & \textbf{4.68} & \textbf{2.16} & \textbf{3.54} & \textbf{1.79} & \textbf{2.54} & \textbf{3.20} & \textbf{4.52} \\
\midrule
\multirow{4}{*}{5L}
  & DFlash & 3.52 & 4.98 & 3.96 & 5.85 & 3.45 & 4.98 & 3.24 & 4.56 & 2.01 & 3.43 & 1.68 & 2.49 & 2.98 & 4.38 \\
  & Top-3 prefix mask & 3.36 & 4.80 & 3.77 & 5.56 & 3.27 & 4.73 & 3.06 & 4.31 & 1.92 & 3.29 & 1.58 & 2.34 & 2.83 & 4.17 \\
  & Accept-rate & 0.74 & 1.03 & 0.73 & 1.03 & 0.73 & 1.02 & 0.74 & 1.02 & 0.71 & 1.01 & 0.74 & 1.01 & 0.73 & 1.02 \\
  & \textbf{D-PACE} & \textbf{3.91} & \textbf{5.54} & \textbf{4.47} & \textbf{6.61} & \textbf{3.81} & \textbf{5.54} & \textbf{3.53} & \textbf{4.99} & \textbf{2.18} & \textbf{3.79} & \textbf{1.73} & \textbf{2.61} & \textbf{3.27} & \textbf{4.85} \\
\midrule
\multicolumn{16}{c}{\emph{Temperature} $T = 1$} \\
\midrule
\multirow{4}{*}{3L}
  & DFlash & 3.23 & 4.37 & 3.41 & 4.91 & 3.16 & 4.37 & 3.04 & 4.11 & 1.93 & 3.05 & 1.67 & 2.32 & 2.74 & 3.86 \\
  & Top-3 prefix mask & 3.13 & 4.23 & 3.20 & 4.59 & 3.04 & 4.19 & 2.88 & 3.88 & 1.85 & 2.94 & 1.60 & 2.25 & 2.62 & 3.68 \\
  & Accept-rate & 0.76 & 1.01 & 0.74 & 1.01 & 0.77 & 1.04 & 0.78 & 1.03 & 0.75 & 1.04 & 0.80 & 1.04 & 0.77 & 1.03 \\
  & \textbf{D-PACE} & \textbf{3.54} & \textbf{4.79} & \textbf{3.70} & \textbf{5.41} & \textbf{3.48} & \textbf{4.83} & \textbf{3.26} & \textbf{4.38} & \textbf{2.04} & \textbf{3.30} & \textbf{1.73} & \textbf{2.45} & \textbf{2.96} & \textbf{4.19} \\
\midrule
\multirow{4}{*}{5L}
  & DFlash & 3.32 & 4.71 & 3.43 & 5.22 & 3.25 & 4.71 & 3.08 & 4.32 & 1.92 & 3.23 & 1.62 & 2.39 & 2.77 & 4.10 \\
  & Top-3 prefix mask & 3.20 & 4.54 & 3.24 & 4.91 & 3.06 & 4.43 & 2.86 & 4.01 & 1.80 & 3.03 & 1.51 & 2.23 & 2.61 & 3.86 \\
  & Accept-rate & 0.74 & 1.03 & 0.73 & 1.03 & 0.73 & 1.02 & 0.74 & 1.02 & 0.70 & 1.01 & 0.74 & 1.01 & 0.73 & 1.02 \\
  & \textbf{D-PACE} & \textbf{3.72} & \textbf{5.31} & \textbf{3.80} & \textbf{5.73} & \textbf{3.58} & \textbf{5.19} & \textbf{3.33} & \textbf{4.73} & \textbf{2.03} & \textbf{3.49} & \textbf{1.71} & \textbf{2.56} & \textbf{3.03} & \textbf{4.50} \\
\bottomrule
\end{tabular}
}
\end{table}

The related baselines reveal two complementary failure modes. Top-3 prefix mask is too coarse: its hard 0/1 mask discards graded position-wise credit. Accept-rate directly optimizes the surrogate, coupling token likelihood with each position's accepted-length contribution in the same gradient path. It also lacks asymmetric smoothing, allowing early low-confidence positions to suppress later signals. D-PACE separates these roles by computing smoothed weights from the surrogate while retaining token-level CE gradients.

\section{Ablations and Additional Analyses}
\label{sec:ablations}

\subsection{Generalization to Other Targets}
\label{sec:cross-model}
To test whether the improvement transfers beyond Qwen3-4B, we repeat the 3L training setup on two additional target models: Llama-3.1-8B-Instruct and Qwen3-8B, using the ShareGPT~\citep{sharegpt2023} dataset and DFlash's per-target default block sizes ($B{=}10$ for Llama-3.1-8B, $B{=}16$ for Qwen3-8B).

D-PACE improves both $\tau$ and SR on every benchmark for both targets (Table~\ref{tab:cross-model}).

\begin{table}[t]
\centering
\caption{\textbf{Generalization to other targets.} Per-benchmark SR / $\tau$ at $T{=}0$ for 3L drafters of Llama-3.1-8B-Instruct and Qwen3-8B (epoch 6).}
\label{tab:cross-model}
\small
\setlength{\tabcolsep}{4pt}
\begin{tabular}{llcccccc}
\toprule
Target & Method & GSM8K & MATH-500 & HumanEval & MBPP & MT-Bench & Alpaca \\
\midrule
\multirow{2}{*}{Llama-3.1-8B}
 & DFlash             & 2.21 / 2.88 & 2.17 / 2.89 & 2.53 / 3.31 & 2.53 / 3.28 & 1.86 / 2.67 & 1.67 / 2.26 \\
 & \textbf{D-PACE}    & \textbf{2.52 / 3.29} & \textbf{2.47 / 3.28} & \textbf{2.82 / 3.69} & \textbf{2.84 / 3.68} & \textbf{2.08 / 3.00} & \textbf{1.83 / 2.49} \\
\midrule
\multirow{2}{*}{Qwen3-8B}
 & DFlash             & 2.85 / 3.85 & 2.64 / 3.62 & 2.41 / 3.28 & 2.40 / 3.22 & 1.91 / 2.86 & 1.71 / 2.40 \\
 & \textbf{D-PACE}    & \textbf{3.25 / 4.40} & \textbf{3.04 / 4.20} & \textbf{2.75 / 3.76} & \textbf{2.65 / 3.56} & \textbf{2.07 / 3.19} & \textbf{1.82 / 2.60} \\
\bottomrule
\end{tabular}
\end{table}

\subsection{Component Ablation}
\label{sec:component-ablation}
To analyze the contribution of each factor, we test two variants: \emph{cumulative-only} ($\bar w_j = \prod_{i\le j}\tilde q_i$) and \emph{continuation-only} ($\bar w_j = \tilde f_j$). Both keep the same smoothing as D-PACE.
Table~\ref{tab:ablation-trajectory} shows two patterns.

\textbf{(1) The continuation value alone gives limited improvement.} The continuation-only variant drops the cumulative factor $\prod_{i\le j}\tilde q_i$ and uses $\tilde f_j$ as the weight. It therefore emphasizes positions with large remaining continuation value, but ignores whether the prefix reaches position $j$ with high probability. After the first few epochs, the variant remains about $2\%$ above DFlash in $\tau$ (Table~\ref{tab:ablation-trajectory}), but stays $7\%$--$9\%$ below cumulative-only and D-PACE.

\textbf{(2) Prefix acceptance provides the dominant signal.} Cumulative-only weights position $j$ by $\prod_{i\le j}\tilde q_i$, a smoothed proxy for the probability of accepting the prefix through position $j$. The cumulative-only variant exceeds DFlash by $10\%$--$12\%$ in $\tau$ at every epoch (Table~\ref{tab:ablation-trajectory}).

Full D-PACE combines both factors: the cumulative term estimates whether the prefix reaches position $j$, while $\tilde f_j$ estimates the remaining accepted-length value from position $j$ onward. This combined weighting yields a smaller but consistent improvement over cumulative-only across epochs (Table~\ref{tab:ablation-trajectory}).

\begin{table}[t]
\caption{\textbf{Component ablation on MATH-500 (3L).} MATH-500 $\tau$ over training epochs for the DFlash baseline, D-PACE, and its two component-ablation variants (continuation-only, cumulative-only).}
\label{tab:ablation-trajectory}
\centering
\small
\begin{tabular}{lccccccccc}
\toprule
Epoch & 1 & 2 & 3 & 4 & 5 & 6 & 7 & 8 & 9 \\
\midrule
DFlash (decayed-CE)    & 4.27          & 4.73          & 5.01          & 5.21          & 5.36          & 5.44          & 5.46          & 5.48          & 5.48 \\
continuation-only      & 4.20          & 4.75          & 5.10          & 5.35          & 5.48          & 5.60          & 5.60          & 5.60          & 5.58 \\
cumulative-only        & 4.75          & 5.26          & 5.62          & 5.83          & 5.95          & 6.00          & 6.05          & 6.06          & 6.04 \\
\textbf{D-PACE (full)} & \textbf{4.76} & \textbf{5.31} & \textbf{5.63} & \textbf{5.86} & \textbf{6.02} & \textbf{6.12} & \textbf{6.15} & \textbf{6.16} & \textbf{6.13} \\
\bottomrule
\end{tabular}
\end{table}

\subsection{Hyperparameter Sensitivity}
\label{sec:hyperparameter-sensitivity}
\label{sec:smoothing-sweep}

\textbf{Smoothing Parameter $\alpha$.}
The smoothing parameter $\alpha$ (Sec.~\ref{sec:smoothing}) sets a lower bound $\tilde q_i \ge \alpha$, preventing the cumulative product in the weight from vanishing. We sweep $\alpha$ on three 3L draft configurations---Qwen3-4B at $B{=}16$ and $B{=}8$, and Llama-3.1-8B at $B{=}10$---and report average $\tau$ at epoch~6 (Table~\ref{tab:smoothing-sweep}). Apart from $\alpha=0$, where the cumulative product vanishes and $\tau$ stalls near $1.0$ across the three configurations, every tested $\alpha$ outperforms the DFlash decayed-CE baseline. At the opposite extreme, $\alpha{=}0.9$ pushes every smoothed confidence near 1 and removes most of the adaptive signal. Performance is stable across $\alpha\in[0.3, 0.7]$, with $\alpha=0.3$ and $\alpha=0.5$ performing comparably.

\begin{wraptable}{r}{0.46\textwidth}
\centering
\vspace{-20pt}
\caption{\textbf{Block size sensitivity.} Average SR and $\tau$ over the six benchmarks on Qwen3-4B (3L) at epoch~6, across three block sizes.}
\label{tab:block-size}
\vspace{0.3em}
\small
\setlength{\tabcolsep}{3pt}
\begin{tabular}{lcccccc}
\toprule
        & \multicolumn{2}{c}{DFlash} & \multicolumn{2}{c}{D-PACE} & \multicolumn{2}{c}{$\Delta$ (\%)} \\
\cmidrule(lr){2-3} \cmidrule(lr){4-5} \cmidrule(lr){6-7}
$B$     & SR & $\tau$ & SR & $\tau$ & SR & $\tau$ \\
\midrule
8       & 2.85 & 3.88 & \textbf{2.97} & \textbf{4.08} & $+4.2$ & $+5.2$ \\
12      & 2.95 & 4.08 & \textbf{3.16} & \textbf{4.42} & $+7.1$ & $+8.3$ \\
16      & 2.93 & 4.10 & \textbf{3.20} & \textbf{4.52} & $+9.2$ & $+10.2$ \\
\bottomrule
\end{tabular}
\vspace{-40pt}
\end{wraptable}

\textbf{Block Size $B$.}
We re-train 3L drafts at $B \in \{8, 12, 16\}$, following DFlash's per-block-size settings. D-PACE improves average SR and $\tau$ over the DFlash baseline at every block size. This trend is consistent with the role of adaptive weighting: longer blocks expose more positions whose contribution depends on prefix acceptance, while D-PACE adjusts weights from the current block (Table~\ref{tab:block-size}).

\begin{table}[t]
\centering
\caption{\textbf{Weight smoothing hyperparameter $\alpha$.} Average $\tau$ over the six benchmarks at epoch~6 across three 3L draft configurations.}
\label{tab:smoothing-sweep}
\small
\setlength{\tabcolsep}{5pt}
\begin{tabular}{lcccccc}
\toprule
                           & \multicolumn{5}{c}{D-PACE ($\alpha$)} & DFlash \\
\cmidrule(lr){2-6} \cmidrule(lr){7-7}
                           & $0.0$ & $0.3$ & $\mathbf{0.5}$ & $0.7$ & $0.9$ & decayed-CE \\
\midrule
Qwen3-4B ($B{=}16$)        & $1.02$ & $4.51$ & $\mathbf{4.52}$ & $4.47$ & $4.32$ & $4.10$ \\
Qwen3-4B ($B{=}8$)         & $1.02$ & $\mathbf{4.09}$ & $4.08$ & $4.04$ & $3.98$ & $3.88$ \\
Llama-3.1-8B ($B{=}10$)    & $1.03$ & $\mathbf{3.26}$ & $3.24$ & $3.18$ & $2.99$ & $2.88$ \\
\bottomrule
\end{tabular}
\end{table}

\begin{wraptable}{r}{0.50\textwidth}
\centering
\vspace{-18pt}

\caption{\textbf{Target-probability surrogate.} Average SR and $\tau$ over the six benchmarks on the 3L draft at epoch~6. \emph{D-PACE (target prob)} replaces the draft probability with the target probability on the same token in the D-PACE weight.}
\label{tab:target}
\small
\begin{tabular}{lcc}
\toprule
Method & SR & $\tau$ \\
\midrule
DFlash (decayed-CE)          & 2.93 & 4.10 \\
D-PACE (target prob)         & 3.01 & 4.25 \\
\textbf{D-PACE (draft prob)} & \textbf{3.20} & \textbf{4.52} \\
\bottomrule
\end{tabular}
\vspace{-15pt}
\end{wraptable}

\subsection{Target-Probability Surrogate}
\label{sec:target-prob-surrogate}

As an analytic variant, we replace the draft-side acceptance proxy $\qpsi(z_i^*)$ in the D-PACE weight with the target probability $\ptgt(z_i^*)$ on the same token. In both SR and $\tau$, this variant remains much closer to the DFlash baseline than to full D-PACE (Table~\ref{tab:target}). Because $\ptgt(z_i^*)$ is independent of the draft model, it cannot capture which positions currently limit acceptance for the given drafter and example.

\subsection{Position-Weight Dynamics}
\label{sec:weight-dynamics}
\emph{Over training}, D-PACE assigns larger weights to low-index block positions in the early training stage, then shifts weight toward higher-index positions as those positions become more likely to limit acceptance (Fig.~\ref{fig:teaser-results}a); in contrast, DFlash's fixed exponential decay remains constant. \emph{At each training step}, D-PACE recomputes weights from the current block, so the position weights vary across examples and updates (Fig.~\ref{fig:teaser-results}b).

\subsection{Training-time overhead}

\begin{wraptable}{r}{0.5\textwidth}
\centering
\vspace{-20pt}
\small
\caption{Training wall-clock on a single H200 for the 3L draft ($B{=}16$). Per-step time averages over 49{,}993 iterations.}
\label{tab:training-time}
\begin{tabular}{lcc}
\toprule
Method & per epoch & per step \\
\midrule
DFlash (decayed-CE) & 2:24 & 173~ms \\
\textbf{D-PACE}     & 2:28 & 177~ms \\
\bottomrule
\end{tabular}
\end{wraptable}

Training-time overhead remains small. D-PACE adds a weight-computation step to each training iteration, but the block-level weights are computed in parallel over all positions. In our measurement on a single H200, this increases per-step time by only 2.3\% compared with the DFlash decayed-CE baseline (Table~\ref{tab:training-time}).

\section{Related Work}
\label{sec:related}

\paragraph{Speculative decoding architectures.}
Speculative decoding proposes tokens with a draft model and verifies them with the target model~\citep{leviathan2023speculative,chen2023speculative}. Prior work has primarily improved speculative decoding through drafter architecture and verification topology. Medusa~\citep{cai2024medusa} and Hydra~\citep{ankner2024hydra} attach multiple prediction heads to target-model features. The EAGLE series predicts at the feature level, adding dynamic tree pruning in EAGLE-2 and training-time test over intermediate features in EAGLE-3~\citep{li2024eagle,li2024eagle2,li2025eagle3}. Other drafters reuse target signals (GLIDE~\citep{du2024glide}, Kangaroo~\citep{liu2024kangaroo}), introduce recurrence (ReDrafter~\citep{cheng2024redrafter}), or use Jacobi-style iteration (Lookahead~\citep{fu2024lookahead}). On the verification side, SpecInfer~\citep{miao2024specinfer} verifies top-$K$ draft trajectories as a tree.

\paragraph{Parallel and diffusion-based drafters.}
Parallel block drafters reduce the sequential dependence on block length by predicting multiple positions in one forward pass~\citep{stern2018blockwise}. This direction connects to diffusion language models, which generate masked positions through parallel denoising steps, including LLaDA~\citep{nie2025llada}, Dream~\citep{ye2025dream}, Mercury~\citep{khanna2025mercury}, and block-diffusion models such as BD3-LM~\citep{arriola2025block}. DFlash~\citep{chen2026dflash} uses a flash-attention block decoder with KV injection from the target's prefill cache; DART~\citep{liu2026dart} predicts multiple future logits and prunes a draft tree; DiffuSpec~\citep{li2025diffuspec} uses a pre-trained dLLM as a training-free drafter with causal-consistency path search. D-PACE focuses on this parallel block-drafting regime, where the training objective determines how learning signal is allocated across positions in the block.

\paragraph{Acceptance-aware drafter objectives.}
Several methods modify drafter training to better match downstream acceptance. GRIFFIN~\citep{hu2025griffin} masks position $j$'s cross-entropy when any earlier target falls outside the draft's top-$K$; we adapt this mask as a baseline (Sec.~\ref{sec:setup}). SpecDiff-2~\citep{sandler2025specdiff2} introduces streak distillation for diffusion drafters, and OPT-Tree~\citep{wang2025opttree} uses expected accepted length as an inference-time objective for draft-tree construction. Concurrent LK losses~\citep{samarin2026lklosses} study acceptance-oriented token-level objectives for speculative sampling. Target-side supervision methods such as HASS, FSPAD, DistillSpec, online speculative decoding, and CORAL~\citep{zhang2025hass,gui2024fspad,zhou2024distillspec,liu2024osd,weng2025coral} use verifier signals or online feedback. Closest to our setting, DFlash trains its parallel drafter with a fixed exponential decay over block positions. DART~\citep{liu2026dart} also uses a fixed exponential position decay, applied to KL losses over parallel future logits. D-PACE replaces these static schedules with weights derived from an accepted-length proxy. The resulting weight is detached from the cross-entropy, with asymmetric smoothing applied only to the weight while the cross-entropy term keeps its standard gradient.

\section{Limitations and Future Work}
\label{sec:limitations}

\paragraph{Scope.}
Our derivation applies to parallel block drafters that generate all draft positions in one forward pass. For sequential drafters, later draft generation depends on earlier draft hidden states, so accumulated draft-state errors can break the fixed-prefix factorization used by our accepted-length surrogate. In addition, we evaluate D-PACE only for lossless speculative decoding. More broadly, contribution-based weighting may be relevant to other parallel block-prediction settings with prefix-truncated utility. Such settings may include multi-token prediction heads and other blockwise sequence-generation objectives, which we do not evaluate here.

\paragraph{Objective extensions.}
Our experiments evaluate D-PACE as a dynamic weighting scheme for token-level cross-entropy training in parallel block drafters such as DFlash. App.~\ref{app:dpakl} (\textbf{D-PAKL}) derives the corresponding KL-distillation variant by replacing the soft acceptance analog $\langle p_i,\qpsi\rangle$ with the Jensen lower-bound proxy $\exp(-C_i)$. Empirical evaluation of D-PAKL and other forms of acceptance-aware dynamic loss weighting remains future work.

\paragraph{Acceptance surrogate.}
D-PACE uses $q_i=q_\psi(z_i^*\mid\text{prefix})$, the draft probability of the token selected by the target decoding policy, as a smooth proxy for whether position $i$ will be accepted. Under hard-match verification, this is not an exact acceptance probability: acceptance is a discrete match event, whereas $q_i$ is the drafter's confidence on the target-selected token. Especially early in training, this confidence proxy may be weakly aligned with realized acceptance; we do not separately evaluate that effect. We rely on the proxy empirically, supported by the correlation between $\tilde S$ and reported $\tau$ in Fig.~\ref{fig:motivation} and App.~\ref{app:correlation}.

\paragraph{Broader impact.}                                    D-PACE improves DFlash's acceleration performance without altering the target model's distribution, translating to lower energy consumption and serving cost at scale. It broadly applies to existing models rather than enabling new capabilities, introducing no novel risks beyond those inherent in the underlying LLMs.

\section{Conclusion}
\label{sec:conclusion}

\textbf{D-PACE} replaces DFlash's fixed exponential decay with per-position weights derived from the gradient of an accepted-length surrogate, tying the drafter's training signal to accepted length, a key factor in speculative decoding efficiency. With 2.3\% training-time overhead (Table~\ref{tab:training-time}) and no architectural or inference-pipeline changes, D-PACE improves average emitted length and speedup on DFlash drafts of Qwen3-4B (reaching up to $4.47\times$ speedup on MATH-500; Table~\ref{tab:main}) and transfers to Llama-3.1-8B and Qwen3-8B (Sec.~\ref{sec:cross-model}). Beyond DFlash, the same construction may apply to other single-trajectory drafters when a differentiable proxy for acceptance is available. More broadly, D-PACE suggests that multi-token drafters can replace fixed position schedules with weights derived from how each position contributes to accepted length.


\clearpage

\appendix

\section{Confidence--acceptance correlation}
\label{app:correlation}

We compute the per-block Spearman rank correlation between two confidence statistics and reported $\tau$ over all blocks from the 3L DFlash baseline ($B{=}16$) on 128 MATH-500 prompts. For a confidence statistic $s_b$ computed on block $b$, we compute
\begin{equation*}
\rho \;=\; \mathrm{corr}\!\left(\operatorname{rank}(s_b),\, \operatorname{rank}(\tau_b)\right).
\end{equation*}
Table~\ref{tab:correlation} compares the surrogate $\tilde S = \sum_{k=1}^{B}\prod_{i\le k} q_i$ used in Sec.~\ref{sec:surrogate} against the plain confidence sum $\sum_i q_i$.

\begin{table}[h]
\centering
\small
\caption{Per-block Spearman rank correlation between reported $\tau_b$ and two block-level confidence statistics on 128 MATH-500 prompts. $\sum_i q_i$ sums per-position draft confidence, while $\tilde S=\sum_{k=1}^{B}\prod_{i\le k}q_i$ sums cumulative prefix-confidence terms, where $k$ indexes the accepted-prefix length. Both correlations have $p < 10^{-10}$.}
\label{tab:correlation}
\begin{tabular}{lc}
\toprule
Per-block statistic & Spearman $\rho$ \\
\midrule
$\sum_i q_i$                                                       & $0.79$ \\
$\tilde S = \sum_{k}\prod_{i\le k} q_i$ (Fig.~\ref{fig:motivation}) & $\mathbf{0.84}$ \\
\bottomrule
\end{tabular}
\end{table}

The cumulative surrogate $\tilde S$ is more predictive of reported $\tau$ than the plain confidence sum.

\section{Loss formulations}
\label{app:losses}

Let $B$ be the block size, $z_i^*$ the target token at position $i$, $q_i = q_\psi(z_i^* \mid \text{prefix})$ the draft's confidence on $z_i^*$, $\tilde q_i = (1-\alpha)q_i + \alpha$ the smoothed confidence (default $\alpha=0.5$), and $\tilde f_j = 1 + \sum_{m=j+1}^{B}\prod_{i=j+1}^{m}\tilde q_i$ the smoothed continuation value (Sec.~\ref{sec:smoothing}).

\subsection{D-PACE}
This is the main D-PACE loss from Sec.~\ref{sec:smoothing}:
\[
\mathcal{L}_{\text{D-PACE}} \;=\; \sum_{j=1}^{B}\,\bar w_j\,(-\log q_j),
\qquad
\bar w_j \;=\; \Bigl(\prod_{i \le j}\tilde q_i\Bigr)\tilde f_j.
\]

\subsection{Baselines}

\paragraph{DFlash decayed-CE (Sec.~\ref{sec:setup}).}
The released DFlash baseline applies an exponential position decay~\citep{chen2026dflash} with a block-size-specific decay constant:
\[
\mathcal{L}_{\text{DFlash}} \;=\; \sum_{j=1}^{B}\,\exp\!\Bigl(-\frac{j-1}{\gamma_B}\Bigr)\,(-\log q_j).
\]
DFlash sets $\gamma_B$ to $7$, $5$, and $4$ for block sizes $16$, $10$, and $8$, respectively; for $B{=}12$, we use $\gamma_{12}=6$.

\paragraph{Top-$K$ prefix mask (Sec.~\ref{sec:setup}).}
Adapted from GRIFFIN~\citep{hu2025griffin}; we use $K{=}3$ and adopt only this mask, not the full GRIFFIN method:
\[
\mathcal{L}_{\text{top-}K} \;=\; \sum_{j=1}^{B} m_j \,(-\log q_j),
\qquad
m_j \;=\; \prod_{i<j}\mathbf{1}\bigl[z_i^* \in \mathrm{Top}_K\,\qpsi(\cdot\mid\text{prefix})\bigr].
\]
Position $j$ contributes only when every earlier target token lies in the draft's top-$K$ predictions.

\paragraph{Accept-rate (Sec.~\ref{sec:setup}).}
This baseline directly ascends the accepted-length surrogate in Eq.~\ref{eq:S}, with no smoothing or detached weighted-CE construction:
\[
\mathcal{L}_{\text{accept-rate}} \;=\; -\sum_{m=1}^{B}\prod_{j=1}^{m} q_j.
\]

\subsection{Ablation variants of D-PACE}

\paragraph{Cumulative-only and continuation-only (Sec.~\ref{sec:component-ablation}).}
Substituting each factor of $\bar w_j$ separately into $\mathcal{L}_{\text{D-PACE}}$:
\begin{align*}
\mathcal{L}_{\text{cumulative-only}} &\;=\; \sum_{j=1}^{B}\,\Bigl(\prod_{i\le j}\tilde q_i\Bigr)\,(-\log q_j),\\[2pt]
\mathcal{L}_{\text{continuation-only}} &\;=\; \sum_{j=1}^{B}\,\tilde f_j\,(-\log q_j).
\end{align*}

\paragraph{Target-probability surrogate (Sec.~\ref{sec:target-prob-surrogate}).}
This variant replaces each $q_\psi(z_i^*)$ inside the weight by the target probability $p_\theta(z_i^*)$, while keeping the cross-entropy on the draft:
\[
\mathcal{L}_{\text{target-prob}} \;=\; \sum_{j=1}^{B}\,\bar w_j^{\text{target}}\,(-\log q_j),
\qquad
\bar w_j^{\text{target}} \;=\; \Bigl(\prod_{i \le j}\tilde p_i\Bigr)\tilde f_j^{\text{target}},
\]
where $\tilde p_i = (1-\alpha)\,p_\theta(z_i^*) + \alpha$ and $\tilde f_j^{\text{target}}$ is the analogous continuation value computed from $\tilde p_i$.

\section{Training details}
\label{app:training}

In addition to the hyperparameters reported in Sec.~\ref{sec:setup}, training uses sequence length~2048, warmup ratio~0.04, gradient clipping at max-norm~1.0, a cosine learning-rate schedule after warmup, and 512 anchor positions per sequence. Within each paired comparison, only the training objective differs; across reported configurations, draft depth, target model, and block size vary as stated in Sec.~\ref{sec:setup}.

\paragraph{Loss formulations.}
All training losses---D-PACE, the DFlash decayed-CE baseline, the Top-$K$ prefix mask, Accept-rate, and the ablation variants---are consolidated in App.~\ref{app:losses}.

\section{Training compute}
\label{app:training-time}

All training runs use a single NVIDIA H200 GPU. A 3L draft run takes approximately 2:25 per epoch and is trained for 6 epochs, giving roughly 15 H200 GPU-hours per experiment setting. Across the main comparisons, baseline runs, and ablations reported in the paper, total training compute is approximately 465 H200 GPU-hours. Table~\ref{tab:training-time} reports the wall-clock comparison used for the overhead estimate.

\section{D-PAKL: KL-distillation variant}
\label{app:dpakl}

For drafters trained with a KL distillation loss against a soft target distribution $p_i := p_{\theta}(\cdot \mid \text{prefix})$ in place of hard cross-entropy on a sampled token $z_i^*$, define the soft-label cross-entropy
\[
C_i \;:=\; -\!\sum_x p_i(x)\,\log \qpsi(x).
\]
The KL loss satisfies $\mathrm{KL}(p_i \,\Vert\, \qpsi)=C_i-H(p_i)$, so $C_i$ and the KL have the same gradient with respect to the draft parameters. The hard-CE setting in Sec.~\ref{sec:method} is the special case $p_i = \delta_{z_i^*}$.

\subsection{Accepted-length surrogate}
\label{app:kl-surrogate}

In Sec.~\ref{sec:surrogate}, the hard-label surrogate uses the draft confidence $q_i = \qpsi(z_i^* \mid \text{prefix})$ as the proxy for the per-position acceptance rate $a_i$ in $\mathbb{E}[X] = \sum_{k=1}^{B}\prod_{i=1}^{k} a_i$. With a soft target, no single sampled token $z_i^*$ exists; the analog of $q_i$ is the inner product $\langle p_i, \qpsi\rangle = \sum_x p_i(x)\,\qpsi(x)$, which collapses to $q_i$ on a Dirac target. By Jensen's inequality on the concave $\log$,
\begin{equation}
\exp(-C_i) \;=\; \exp\!\bigl(\mathbb{E}_{p_i}[\log \qpsi]\bigr) \;\le\; \mathbb{E}_{p_i}[\qpsi] \;=\; \langle p_i, \qpsi\rangle,
\label{eq:dpakl-jensen}
\end{equation}
with equality on a Dirac target. We use the Jensen lower bound as the proxy,
\begin{equation}
\hat a_i \;:=\; \exp\!\bigl(-C_i\bigr),
\label{eq:dpakl-proxy}
\end{equation}
so that $-\log \hat a_i = C_i$ by construction, and $\hat a_i$ requires no extra forward computation. The resulting surrogate is
\begin{equation}
\tilde S(\psi) \;:=\; \sum_{k=1}^{B}\prod_{i=1}^{k}\hat a_i.
\label{eq:dpakl-Stilde}
\end{equation}

\subsection{Per-position coefficient}
\label{app:kl-coef}

Differentiating Eq.~\eqref{eq:dpakl-Stilde} with respect to each $\hat a_j$ as in Sec.~\ref{sec:coefficient}, split the outer sum at $k=j$:
\[
\tilde S \;=\; \sum_{k=1}^{j-1}\prod_{i=1}^{k}\hat a_i \;+\; \Bigl(\prod_{i<j}\hat a_i\Bigr)\, \hat a_j \,\Bigl(1 + \sum_{m=j+1}^{B}\prod_{i=j+1}^{m}\hat a_i\Bigr) \;=\; \sum_{k=1}^{j-1}\prod_{i=1}^{k}\hat a_i \;+\; \Bigl(\prod_{i<j}\hat a_i\Bigr)\, \hat a_j\, f_j,
\]
where the continuation value is
\[
f_j \;:=\; 1 + \sum_{m=j+1}^{B}\prod_{i=j+1}^{m}\hat a_i.
\]
The first sum is constant in $\hat a_j$, so $\partial \tilde S / \partial \hat a_j = (\prod_{i<j}\hat a_i)\, f_j$. Applying $\nabla_\psi \hat a_j = \hat a_j\, \nabla_\psi \log \hat a_j$ and using $-\log \hat a_j = C_j$,
\begin{equation}
\nabla_\psi \tilde S \;=\; \sum_{j=1}^{B}\Bigl(\prod_{i\le j}\hat a_i\Bigr)\, f_j\, \nabla_\psi \log \hat a_j \;=\; -\sum_{j=1}^{B} w_j\, \nabla_\psi C_j, \qquad w_j \;:=\; \Bigl(\prod_{i\le j}\hat a_i\Bigr)\, f_j.
\label{eq:dpakl-grad}
\end{equation}

\subsection{Weight smoothing}
\label{app:kl-smoothing}

The cumulative product $\prod_{i\le j}\hat a_i$ vanishes when several $\hat a_i$ are small. Following Sec.~\ref{sec:smoothing}, we smooth each $\hat a_i$ inside the weight only, leaving $C_j$ unsmoothed:
\[
\tilde a_i \;:=\; (1-\alpha)\,\hat a_i + \alpha,
\qquad
\tilde f_j \;:=\; 1 + \sum_{m=j+1}^{B}\prod_{i=j+1}^{m}\tilde a_i,
\qquad
\bar w_j \;:=\; \Bigl(\prod_{i \le j}\tilde a_i\Bigr)\, \tilde f_j.
\]

\subsection{D-PAKL loss}
\label{app:kl-norm}

Substituting $\bar w_j$ as the per-position weight gives the D-PAKL loss
\[
\mathcal{L}_{\text{D-PAKL}}(\psi) \;=\; \sum_{j=1}^{B} \bar w_j\, C_j,
\]
with $\bar w_j$ detached from the gradient. Equivalently, one may replace $C_j$ by $\mathrm{KL}(p_j\Vert\qpsi)$ in the final weighted loss because $H(p_j)$ is constant with respect to the draft parameters. In the hard-CE limit, Eq.~\eqref{eq:dpakl-jensen} is tight, $\hat a_i = q_i$, $C_j = -\log q_j$, and D-PAKL reduces to D-PACE (Eq.~\ref{eq:dpace-loss}).

\end{document}